\documentclass[letterpaper, 10 pt, conference]{ieeeconf}  

\IEEEoverridecommandlockouts                              

\usepackage{graphics} 
\usepackage{epsfig} 
\usepackage{amsmath} 
\usepackage{amssymb}  
\usepackage[ruled]{algorithm2e}
\usepackage{tabularx}
\usepackage[table]{xcolor}
\usepackage{hhline}
\usepackage{multirow}
\usepackage{listings}
\usepackage{color}
\usepackage{url}
\usepackage{hyperref}
\usepackage{scrextend}

\newcommand\CC{C\nolinebreak[4]\hspace{-.05em}\raisebox{.45ex}{\relsize{-3}{\textbf{++}}}}

\title{\LARGE \bf
Interactive Co-Design of Form and Function for Legged Robots using the Adjoint Method
}

\author{\authorblockN{Ruta Desai, Beichen Li, Ye Yuan and Stelian Coros}
\thanks{This work was supported by the National Science Foundation. The authors are with the Robotics Institute, Carnegie Mellon University, USA. {\tt\footnotesize \{rutad, beichen1, yyuan2, scoros\}@andrew.cmu.edu}}
}

\begin{document}

\maketitle
\thispagestyle{empty}
\pagestyle{empty}

\begin{abstract}
Our goal is to make robotics more accessible to casual users by reducing the domain knowledge required in designing and building robots. Towards this goal, we present an interactive computational design system that enables users to design legged robots with desired morphologies and behaviors by specifying higher level descriptions. The core of our method is a design optimization technique that reasons about the structure, and motion of a robot in coupled manner in order to achieve user-specified robot behavior, and performance. We are inspired by the recent works that also aim to jointly optimize robot's form and function. However, through efficient computation of necessary design changes, our approach enables us to keep user-in-the-loop for interactive applications. We evaluate our system in simulation by automatically improving robot designs for multiple scenarios. Starting with initial user designs that are physically infeasible or inadequate to perform the user-desired task, we show optimized designs that achieve user-specifications, all while ensuring an interactive design flow.

\end{abstract}

\section{Introduction}
\label{sec:intro}
Even from a cursory inspection, it is clear that the morphological features of living creatures are intimately related to their motor capabilities. The long limbs and flexible spine of a cheetah, for instance, lead to dizzying speeds. When designing robots for different tasks, it is therefore not surprising that roboticists often look to nature for inspiration. Examples include robots that mimic salamanders~\cite{crespi2013salamandra}, kangaroos~\cite{graichen2015control}, and even insects~\cite{altendorfer2001rhex}. However, rather than copying the designs that we see in nature, we are interested in beginning to address the question: can we develop mathematically-rigorous models with the predictive power to inform the design of effective legged robots? Furthermore, we are interested in integrating such computational models into interactive design tools that make robotics accessible to casual users.

Towards this goal, we develop a computationally efficient interactive design system that allows users to create legged robots with diverse functionalities, without requiring any domain-specific knowledge. Our system automatically suggests required changes to a user-designed robot morphology in order to achieve a specified behavior or task performance. In particular, we focus on periodic locomotion-based tasks that are characterized by footfall patterns, walking/turning speed, and direction of motion. Our approach is therefore a departure from conventional, largely manual trial and error approaches that iteratively improve task-based robot designs. The core of our system consists of a mathematical model that maps the morphological parameters of a robot to its motor capabilities. Equipped with this model that captures the sensitivity of robot's motion on its morphology, we present an automatic design framework to co-optimize them in a hierarchical manner.
To deal with the computational complexity, and to enable user interactivity throughout the design process, we integrate the highly scalable Adjoint method~\cite{giles2000introduction} within our framework. 

Finally, we validate our system in a physically-simulated environment, through various task-based robot design scenarios that are challenging for casual users. We demonstrate how our system is able to aid users in dealing with a variety of such issues ranging from physical in-feasibility of the design, to sub-optimality in task performance. 
\section{Related Work}
\label{sec:relatedwork}

Our long term goal is to make the process of creating customized robots highly accessible. We share this goal with many others in the robotics community~\cite{schulz2017interactive, bezzo2015robot,Mehta:2014:IROS}. 
Towards our goal, in the past, we developed an interactive design system for rapid, and on-demand generation of custom robotic devices~\cite{desai2017computational}. Our system was an intuitive forward design tool that allowed novices to explore the design space of robotic devices, and to test their designs in simulations. 
However, when a user-designed robot failed to match the users' desired behavior or task performance, our system did not provide any feedback to the users about improving their designs. Instead, we relied completely on the users' intuition and experience to improve their designs, leveraging real-time feedback from the physical simulation.

Recently, various approaches have been proposed to either provide feedback to the users about improving their designs~\cite{canaday2017interactive}, or for automatically changing them based on a desired outcome~\cite{hajoint,spielberg2017functional}. We are highly inspired by these approaches that leverage the coupling between the robot's morphology and behavior. In particular, Canaday \emph{et al.} and Ha \emph{et al.} update the robot's morphology for a given behavior such that kinematic and actuator constraints are satisfied~\cite{canaday2017interactive,hajoint}. On the contrary, we co-optimize the morphology and motion of a robot without assuming a predefined motion or control policy for improving the robot's performance in locomotion-based tasks, similar to~\cite{spielberg2017functional}. However, our formulation is much more efficient than~\cite{spielberg2017functional}, making it well-suited for user-in-the-loop optimization. Unlike~\cite{spielberg2017functional}'s approach of directly weaving in the robot's physical parameters within the robot's motion optimization framework, we establish a mapping between the robot's physical and motion parameters using the sensitivity analysis or implicit function theorem~\cite{jittorntrum1978implicit}. Ha \emph{et al.} also use such a mapping between robot's form and function for optimizing robot designs that use minimal actuator forces~\cite{hajoint}. However, they rely on the users to provide a change direction for modifying their designs. Instead, we enable our users to define their design requirements at task level using much wider variety of specifications such as desired speeds, desired directions of motion etc. We also allow the users to optimize their designs for multiple tasks, thereby automating the painstaking process of achieving optimal trade-off between possibly contradicting task requirements.



Such task-specific robot design from high-level descriptions has been a long-standing interest for the robotics community. In particular, a large number of approaches based on evolutionary algorithms have been successfully used for synthesizing morphologies and controllers for a variety of robots such as virtual creatures, manipulators etc.~\cite{Sims94,Leger_1999_3269,EPFL-CONF-200995}. Unfortunately, despite promising early results and significant increases in computing power, the field of evolutionary robotics has shown a limited ability to go beyond Sims' initial work~\cite{Sims94} in terms of complexity and behavioral sophistication~\cite{cheney2016difficulty}. Stochastic approaches also provide limited theoretical guarantees, and are susceptible to produce designs that are often not reproducible. On the other hand, our gradient based approach is locally optimal. Furthermore, rather than synthesizing designs from scratch, we enable the users to specify initial designs, as well as to modify their designs at any point during the design process. Having the user-in-the-loop allows the users to guide the design process as they desire, thereby converging onto their needed outcome much faster. Letting the users to play an active role in creating robotic devices for personal use is also essential for improving the quality of their interactions with the final product~\cite{sun2016psychological}. 


\section{Interactive Design}
\label{sec:method}
\begin{figure}[htbp]
  	\centering
  	\includegraphics[width=\columnwidth]{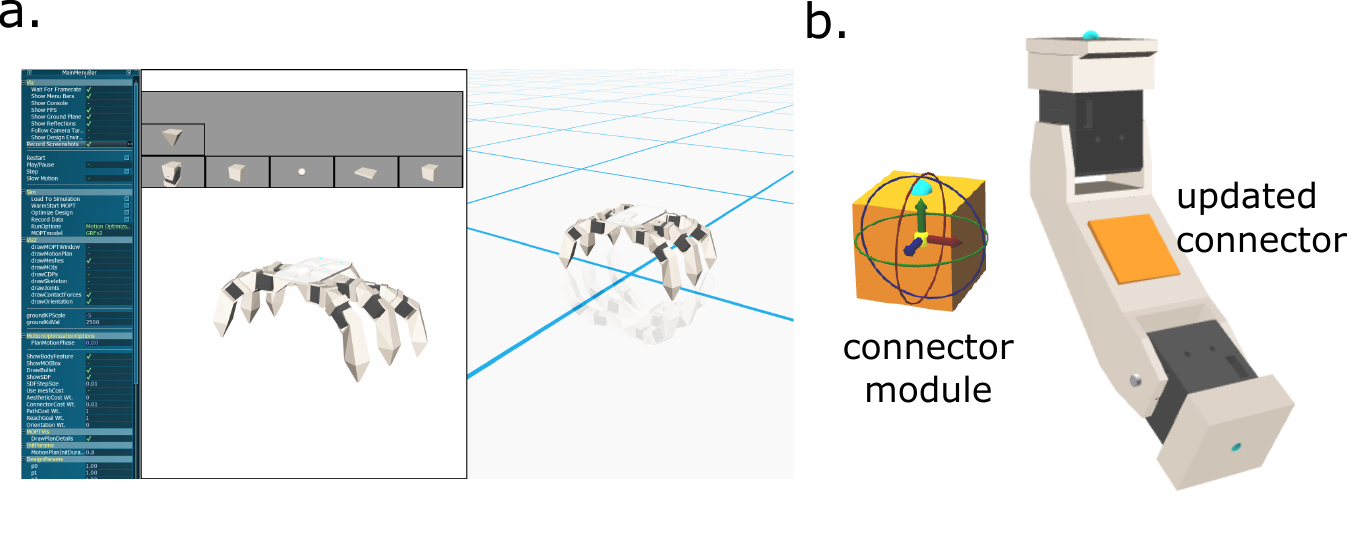}
  	\caption{(a)~Our design interface is adopted from~\cite{desai2017computational}. It allows users to design robots using modular 3D printed, and off-the-shelf parts, as well as test them using physical simulation.~(b)~We use a parameterized 3D printable connector module to automatically create connections between actuators. The configuration, and size of the original connector (shown in orange) gets updated based on the configuration of two actuators that it connects.}
  	\label{gui}
\end{figure}

Our interactive design system is rooted in a design abstraction, that allows us to combine off-the-shelf components such as actuators, and 3D printed parts such as brackets for designing robots~\cite{desai2017computational}. As figure~\ref{gui}(a) illustrates, our graphical user interface (GUI) consists of a design workspace (left) and a simulation workspace (right). The design workspace allows them to browse through the list of available modules such as actuators and 3D printed parts, which can be dragged and dropped into the scene, to construct and modify a robot design. We have found this to be a powerful approach to specify initial robot morphology, and design. Our video\footnote{\label{video}Video is available at: \url{https://tinyurl.com/RoboCodesign}} demonstrates the design process for one of our examples. Users can also specify and physically simulate robot behaviors (such as walk forward at a desired speed) for their designs in the workspace on the right.

While the GUI is adopted from our previous system~\cite{desai2017computational}, we introduce some design features that better suit our current application. First, to enable the design of more organic looking robots that are optimized for a specific task, and are less cumbersome to assemble, we take a departure from using off-the-shelf brackets for connecting actuators. Instead, we assume that all parts of the robot's articulated structure (other than actuators for joints) are 3D printed. Second, we automatically create these 3D printable connectors between actuators. We are inspired by the robots created in~\cite{megaro2015} that create convex hull geometries between actuators. However, while~\cite{megaro2015} create these geometries as a final processing step, we enable the connector geometry update during interactive design. With every drag-and-drop user operation that changes the robot morphology, as well as during automatic design optimization, the corresponding connector geometries are updated as well.

To enable this, we define a parameterized 3D printable connector module (see fig.~\ref{gui}(b)). Parameterizing the connecting structure as a module allows us to update its position and orientation interactively with changes in the design. Each connector module is also endowed with `virtual' attachment points on the connector's face, that get updated based on its position as well as that of the actuators it connects to. These attachment points are used to update the shape and size of the module's convex hull structure as needed. Our video\textsuperscript{\ref{video}} demonstrates how the 3D printable connecting structures of a robot's legs change as the design optimization updates the robot's design for a specific task. This allows the users to visualize the design changes as the optimization progresses. If the users disapprove of the aesthetic appearance of the robot at any point in the optimization, they can pause the optimization, and make the necessary changes, before continuing the optimization.

\begin{figure*}[htbp]
     	\centering
     	\includegraphics[width=\textwidth]{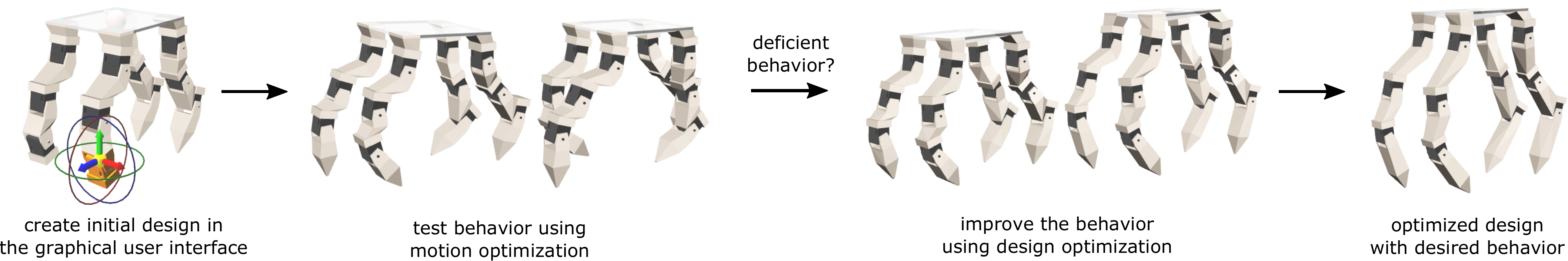}
     	\caption{Overview of the design process: A design session with our system typically begins with users creating initial robot morphologies using our interactive GUI. The behavior of their initial designs can be tested in the physical simulation using our motion optimization framework~(Sec.~\ref{sec:mopt}.) If the robot behavior is not as desired or deficient, instead of manually changing the design, users can use our automatic design optimization~(Sec.~\ref{sec:dopt}) that improves their initial designs for the task at hand. In the example shown, the initial robot couldn't walk as fast as desired. The design optimization lengthens the robot limbs to enable this.}
   	     \label{fig:teaser}
\end{figure*}

Although our interactive design interface is a powerful approach for forward design, modifying designs to achieve desired capabilities for a task requires domain knowledge. To aid novices without such knowledge in creating robots, as well as to enable experts in fine-tuning their robots for specific tasks, we next present our design optimization framework, which automatically optimizes the robot's form and behavior for performing a desired task. In scenarios where the users are not satisfied with the behavior of their initial designs, they can use such an optimization to improve their designs. Fig.~\ref{fig:teaser} illustrates this design process for creating a fast walking quadruped.

\section{Automatic Design Optimization}
Designing robots with task-specific behaviors is highly skill-intensive and time-consuming. One must decide the robot's structure -- physical dimensions of its body, and its articulated parts, as well as the placement of actuators. One must then define how to control the actuators for a co-ordinated movement that achieves a task. The robot's structure has a huge effect on the tasks it can perform. Therefore, designers typically iterate back and forth between physical and motion design to create a task-specific robot. To capture this coupling between the robot's form and function, we parameterize a robot with a set of structure parameters $\mathbf{s}$, and motion parameters $\mathbf{m}$. However, instead of treating $\mathbf{m}$ and $\mathbf{s}$ independently, our goal is to represent robot motions as a function of its structure $\mathbf{m}(\mathbf{s})$. Apart from being intuitive, such a representation allows us to solve for an optimal task-specific behavior and design hierarchically, in a computationally efficient manner. This, in turn, allows us to generate results much faster, enabling interactivity during design. We explain our formulation in detail in this section.



\subsection{Parameterization}
A larger variety of robots including manipulators, and walking robots are composed of articulated chain like structures, in particular, of serially connected and actuated links. Such robot morphologies can be well described as kinematic trees starting at the root of the robot. The design parameters $\mathbf{s}$ is used to specify the robot morphology, which is given by \begin{align}
\mathbf{s} &= \left(l_1, \ldots, l_g, \mathbf{a}_1, \ldots, \mathbf{a}_n, b_w, b_l\right)\,,
\label{eq:robot}
\end{align}
where $g$ is the number of links, $l_i \in \mathbb{R}$ is the length of each link, $n$ is the number of actuators, and $\mathbf{a}_i \in \mathbb{R}^3$  is the actuator parameters. For linear actuators, $\mathbf{a}_i$ defines the 3D attachment points, while for rotary actuators, it corresponds to orientation of axis of rotation. Apart from these parameters that represent the kinematic tree morphology of the robot, we use two additional parameters $b_w$ and $b_l$ to represent the physical dimensions of the robot's body (width and length respectively).


Likewise, the motion parameters $\mathbf{m} = \left( \mathbf{P}_1, \ldots, \mathbf{P}_T \right)$ are defined by a time-indexed sequence of vectors $\mathbf{P}_i$, where $T$ denotes the time for each motion cycle. $\mathbf{P}_i$ is defined as:
\begin{align}
\mathbf{P}_i &= \left(\mathbf{q}_i, \mathbf{x}_i, \mathbf{e}_i^1, \ldots, \mathbf{e}_i^k, \mathbf{f}_i^1, \ldots, \mathbf{f}_i^k, c_i^1, \ldots, c_i^k,\right)\,,
\label{eq:robot2}
\end{align}
where $\mathbf{q}_i$ defines the pose of the robot, i.e., the position, and orientation of the root as well as joint information such as angle values, $\mathbf{x}_i \in \mathbb{R}^3$ is the position of the robot's center of mass (COM), and $k$ is the number of end-effectors. For each end-effector $j$, we use $\mathbf{e}_i^j \in \mathbb{R}^3$ to represent its position and $\mathbf{f}_i^j \in \mathbb{R}^3$ to denote the ground reaction force acting on it. We also use a contact flag $c_i^j$ to indicate whether it should be grounded ($c_i^j=1$) or not ($c_i^j=0$). We note that $\mathbf{s}$ remains invariant over the entire motion optimization horizon.

\subsection{Method Overview}
\label{sec:dopt}
Given an initial robot design, and a task specification, our goal is to change $\mathbf{s}$ and $\mathbf{m}$ (as defined in eq.~\ref{eq:robot},~\ref{eq:robot2}) to obtain a design better suited for a task. Users typically define the initial design using our graphical interface. Various task descriptions such as preferred direction of motion/action, desired movement speed, movement styles (walking, trotting, turning) etc. can also be easily specified using our interface. These task specifications can then be encoded into an evaluation criteria or cost function $F(\mathbf{s},\mathbf{m})$. Assuming $\mathbf{p}$ to be the parameter vector containing both structure and motion parameters~$\mathbf{p}=[\mathbf{s},\mathbf{m}]$, one can search for an optimal $\mathbf{p}$ along the direction of $F(\mathbf{p})$'s gradient $\frac{\partial{F}}{\partial{\mathbf{p}}}$. However, $\mathbf{s}$ and $\mathbf{m}$ are inherently coupled. Therefore, instead of searching $\mathbf{s}$ and $\mathbf{m}$ independently, we adopt a hierarchical approach, wherein we first update $\mathbf{s}$, and then update $\mathbf{m}$ within a constrained manifold that maintains the validity and optimality of $\mathbf{m}$'s update, given $\mathbf{s}$.
 
By constructing a manifold of structure and motion parameters of a robot design, we can explore the sensitivity of robot's motion $\mathbf{m}$ to its structure $\mathbf{s}$. Starting with an initial design $(\mathbf{s_0},\mathbf{m_0})$ on the manifold, one can search for $\mathbf{s}$, and corresponding $\mathbf{m}(\mathbf{s})$ on this manifold, such that $F(\mathbf{s},\mathbf{m})$ is minimized. 
This dependency of $\mathbf{m}$ on $\mathbf{s}$ is captured by the Jacobian $\frac{d\mathbf{m}}{d\mathbf{s}}$ (more details in Sec.~\ref{sec:manifold}). This Jacobian is used to compute the search direction $\frac{dF}{d\mathbf{s}}$ for updating $\mathbf{s}$ within the manifold. However, $\frac{d\mathbf{m}}{d\mathbf{s}}$ is expensive to compute. Therefore, we further simplify this computation by using the Adjoint method (see Sec.~\ref{sec:adjoint}). Algorithm~\ref{algo} succinctly describes these steps. Note that $\delta_1$ and $\delta_2$ are the step sizes of the update of $\mathbf{s}$ and $\mathbf{m}$ respectively, in the desired search directions. The \emph{update($\Delta_{\mathbf{s}}$)} function essentially describes the line-search procedure for $\delta_1$ along the search direction $\Delta_{\mathbf{s}}$, and the corresponding update of $\mathbf{s_i}$ at each iteration $i$. $\Delta_{\mathbf{s}}$ represents the updated search direction for $\mathbf{s}$ obtained using the Limited-memory BFGS algorithm~\cite{liu1989limited}, given the gradient direction $\frac{dF}{d\mathbf{s}}$. $N$ represents the number of maximum line search iterations. $\delta_2$ is also computed using similar back-tracking line-search~\cite{nocedal2006numerical}. For each update of $\mathbf{s_i}$, multiple updates of $\mathbf{m}$ are executed to obtain the corresponding optimal $\mathbf{m_i}$. $\mathbf{m}$ is updated in the search direction defined by Newton's method wherein $\frac{\partial^2{F}}{\partial{\mathbf{m}}^2}$, and $\frac{\partial{F}}{\partial{\mathbf{m}}}$ represent the Hessian and gradient of $F$ with respect to $\mathbf{m}$ respectively. 

\begin{algorithm}
	\DontPrintSemicolon
	\SetAlgoNoLine
	\SetKwInOut{Input}{input}\SetKwInOut{Output}{output}
	\Input{Initial robot design $R$ defined with $\mathbf{s_0}$, and $\mathbf{m_0}$; Task specification $F(\mathbf{s},\mathbf{m})$}
	\Output{$(\mathbf{s}^*, \mathbf{m}^*) \mid F(\mathbf{s}^*, \mathbf{m}^*)<threshold$}
	\SetKwFunction{update}{update}
	\While{$F(\mathbf{s_i},\mathbf{m_i}) > threshold$}{
		compute $\frac{dF}{d\mathbf{s}} \Bigr|_{(\mathbf{s_i},\mathbf{m_i})}$ using the Adjoint method\;
		update $\Delta_{\mathbf{s}}$ with$\frac{dF}{d\mathbf{s}}$ using L-BFGS algorithm\;
		update $(\mathbf{s_i},\mathbf{m_i})$ using \update($\Delta_{\mathbf{s}}$)\; 
	}
	\SetKwProg{Fn}{Function}{ }{end}
	\Fn{update($\Delta_{\mathbf{s}}$)}{
		$\delta_1 = 1$\;
		\For{N}{
		$\mathbf{s}' = \mathbf{s_i} -\delta_1  \Delta_{\mathbf{s}}$\;
		$\mathbf{m}' = \mathbf{m_i} -\delta_2 \left(\frac{\partial^2{F}}{\partial{\mathbf{m}}^2}\right)^{-1}\frac{\partial{F}}{\partial{\mathbf{m}}} $ (Newton's method)\;
		\uIf{$F(\mathbf{s}',\mathbf{m}') < F(\mathbf{s_i},\mathbf{m_i})$}{
		$\mathbf{m_i} = \mathbf{m}', \mathbf{s_i} = \mathbf{s}'$\;
		return ($\mathbf{s_i},\mathbf{m_i})$\;
	}
		\Else{
		$\delta_1 = \frac{\delta_1}{2}$
	}
	}
	return ($\mathbf{s_i},\mathbf{m_i})$\;
	}
	\caption{Automatic design optimization}
	\label{algo}
\end{algorithm}

\subsection{Coupling form and function for robot design}
\label{sec:manifold}
It is hard to analytically represent the dependency or sensitivity of robot's motion on its structure. Instead, we assume a manifold that relates robot's structure and behavior capabilities, given a specific task.
\begin{align}
\mathbf{G}(\mathbf{s},\mathbf{m}) = 0,
\label{eqn:manifold}
\end{align}
where $\mathbf{G}(\mathbf{s},\mathbf{m}): \mathbb{R}^{n_s} \times \mathbb{R}^{n_m} \rightarrow \mathbb{R}^{n_m}$. It allows us to understand the effect of design parameters on the motion/behavior of the robot. Such an implicit manifold between structure and function can be converted into an explicit relation between the two within a small region around a point $P_0(\mathbf{s_0},\mathbf{m_0})$ on the manifold, using Implicit function theorem~\cite{jittorntrum1978implicit}. The theorem states that when we change $\mathbf{s_0}$ and $\mathbf{m_0}$ by $\Delta{\mathbf{s}}$ and $\Delta{\mathbf{m}}$, the change in the function $\Delta{\mathbf{G}}$ should be zero to remain on the manifold. Using chain rule to compute $\Delta{\mathbf{G}}$, we obtain the following explicit relation between $\Delta{\mathbf{s}}$~and~$\Delta{\mathbf{m}}$. 
\begin{align}
\Delta{\mathbf{G}} &= \frac{\partial{\mathbf{G}}}{\partial{\mathbf{s}}} \Delta{\mathbf{s}} + \frac{\partial{\mathbf{G}}}{\partial{\mathbf{m}}} \Delta{\mathbf{m}} = 0\,\nonumber\\
\Delta{\mathbf{m}} &= -\left(\frac{\partial{\mathbf{G}}}{\partial{\mathbf{m}}}\right)^{-1}\frac{\partial{\mathbf{G}}}{\partial{\mathbf{s}}} \Delta{\mathbf{s}}\,
\label{eq:jacobian}
\end{align}
where $\left(\frac{\partial{\mathbf{G}}}{\partial{\mathbf{m}}}\right)$, and $\left(\frac{\partial{\mathbf{G}}}{\partial{\mathbf{s}}}\right)$ represents the Jacobian of $\mathbf{G}(\mathbf{s},\mathbf{m})$ with respect to $\mathbf{m}$, and $\mathbf{s}$ respectively. 

In order to compute such a manifold, we start with a task-specific cost function $F(\mathbf{s},\mathbf{m})$. For each robot morphology defined by $\mathbf{s}$, there exists an optimal $\mathbf{m}^*$ that minimizes $F(\mathbf{s},\mathbf{m})$. Therefore, the gradient of $F$ with respect to $\mathbf{m}$ at point $(\mathbf{s}, \mathbf{m}^*)$ should be zero. One can then search for an optimal $\mathbf{s}^*$ along the manifold defined by this gradient $\mathbf{G}(\mathbf{s},\mathbf{m})~=~\frac{\partial{F(\mathbf{s},\mathbf{m})}}{\partial{\mathbf{m}}}$. An optimal $\mathbf{s}^*$ on such a $\mathbf{G}(\mathbf{s},\mathbf{m})$ would automatically ensure a corresponding valid and optimal $\mathbf{m}^*$ for the task. 

For searching such an optimal $\mathbf{s}^*$, we therefore need to solve the following optimization problem:
\begin{align}
\min_{s} F(\mathbf{s},\mathbf{m}) \, \nonumber \\
s.t. \quad \mathbf{G}(\mathbf{s},\mathbf{m}) = 0 \,
\label{eq:designOpt}
\end{align}
where $F(\mathbf{s},\mathbf{m}): \mathbb{R}^{n_s} \times \mathbb{R}^{n_m} \rightarrow \mathbb{R}$ is the energy function; $\mathbf{G}(\mathbf{s},\mathbf{m}): \mathbb{R}^{n_s} \times \mathbb{R}^{n_m} \rightarrow \mathbb{R}^{n_m}$ denotes the gradient of energy function with respect to motion parameters $\mathbf{m}$ and thus $\mathbf{G}~=~ \frac{\partial{F}}{\partial{\mathbf{m}}}$. Empowered by the Jacobian $\frac{d\mathbf{m}}{d\mathbf{s}}$ that essentially encodes $\mathbf{m}(\mathbf{s})$ (defined by eq.~\ref{eq:jacobian}), we can define the search direction for $\mathbf{s}$ as follows:

\begin{align}
\frac{dF}{d\mathbf{s}} &= \frac{\partial{F}}{\partial{\mathbf{m}}} \frac{d\mathbf{m}}{d\mathbf{s}} + \frac{\partial{F}}{\partial{\mathbf{s}}} \nonumber \\
&= - \frac{\partial{F}}{\partial{\mathbf{m}}} \left(\frac{\partial{\mathbf{G}}}{\partial{\mathbf{m}}}\right)^{-1} \frac{\partial{\mathbf{G}}}{\partial{\mathbf{s}}} + \frac{\partial{F}}{\partial{\mathbf{s}}}.
\label{eq:searchs}
\end{align}  

\subsection{The Adjoint method}
\label{sec:adjoint}

Computing $\frac{dF}{d\mathbf{s}}$ requires the calculation of the Jacobian $\frac{d\mathbf{m}}{d\mathbf{s}}$ which is computationally very expensive. It requires solving $n_s$ linear equations (for each column in Jacobian matrix $\frac{\partial{\mathbf{G}}}{\partial{\mathbf{s}}}$), and the procedure gets very costly for large $n_s$. Instead, we use the Adjoint method to efficiently compute the gradient $\frac{dF}{d\mathbf{s}}$. This method formulates the computation of gradient as  constrained optimization problem, and then uses the dual form of this optimization problem for faster computation~\cite{giles2000introduction}. Other applications have also sought out the Adjoint method for similar purposes in the past~\cite{mcnamara2004fluid}.



In particular, the expression $- \frac{\partial{F}}{\partial{\mathbf{m}}} \left(\frac{\partial{\mathbf{G}}}{\partial{\mathbf{m}}}\right)^{-1}$ in eq.~\ref{eq:searchs} can be interpreted as the solution to the linear equation
\begin{align}
\left(\frac{\partial{\mathbf{G}}}{\partial{\mathbf{m}}}\right)^\intercal \lambda = -\left(\frac{\partial{F}}{\partial{\mathbf{m}}}\right)^\intercal \,,
\label{eq:adjoint4}
\end{align}
where $^\intercal$ is the matrix transpose operator. The vector $\lambda$ is called the vector of \textit{adjoint variables} and the linear equation is called the \textit{adjoint equation}. Finally, $\frac{dF}{d\mathbf{s}}$ takes on the following form
\begin{align}
\frac{dF}{d\mathbf{s}} = \lambda^\intercal \frac{\partial{\mathbf{G}}}{\partial{\mathbf{s}}} + \frac{\partial{F}}{\partial{\mathbf{s}}} \,.
\label{eq:adjoint5}
\end{align}

Such a computation of $\frac{dF}{d\mathbf{s}}$ now involves solving only one linear equation (eq.~\ref{eq:adjoint4}), followed by one matrix-vector multiplication and one vector addition (eq.~\ref{eq:adjoint5}). This is much more efficient as compared to solving $n_s$ linear equations for $\frac{d\mathbf{m}}{d\mathbf{s}}$ computation earlier.


\subsection{Motion Optimization}
\label{sec:mopt}
So far, we have described our framework to optimize the structure and motion of a robot, given a task. We used a cost function $F(\mathbf{s},\mathbf{m})$ to encode the task specifications. We now describe how $F(\mathbf{s},\mathbf{m})$ is constructed. To this end, we use a set of objectives that capture users' requirements, and constraints that ensure task feasibility. 

\subsubsection{Objectives}
We allow the users to define various high-level goals to be achieved by their robot designs such as moving in desired direction with specific speeds, different motion styles, etc. To capture the desired direction and speed of motion, we define the following objectives:

\begin{align}
E_{\text{Travel}} &= \frac{1}{2}||\mathbf{x}_T - \mathbf{x}_1 - \mathbf{d}^D||^2 \,,\nonumber \\
E_{\text{Turn}} &= \frac{1}{2}||\tau(\mathbf{q}_T) - \tau(\mathbf{q}_1) - \tau^D||^2\,,
\label{eq:motionObj1}
\end{align}
where $\mathbf{x_i}$ is the robot's COM as defined in eq.~\ref{eq:robot2}, $\tau(\mathbf{q_i})$ is the turning angle computed from pose $\mathbf{q_i}$, while $\mathbf{d}^D$ and $\tau^D$ are desired travel distance and turning angles respectively. $E_{\text{Travel}}$ ensures that the robot travels a specific distance in desired time, while $E_{\text{Turn}}$ can be used to make a robot move on arbitrary shaped paths. 

Motion style is highly effected by gait or foot-fall patterns that define the order and timings of individual limbs of a robot. We internally define various foot-fall patterns for different motion styles such as trotting, pacing, and galloping. When users select a specific motion style, our system automatically loads the necessary foot-fall patterns, thereby allowing novice users to create many expressive robot motions. Motion style is also effected by robot poses. For expert users, we allow the capability to specify and achieve desired poses, if needed, using the following objectives:
\begin{align}
E_{\text{StyleCOM}} &= \frac{1}{2}\sum_i^T||\mathbf{x}_i - \mathbf{x}_i^D||^2 \,,\nonumber \\
E_{\text{StyleEE}} &= \frac{1}{2}\sum_i^T \sum_j^k||\mathbf{e}_i^j - {\mathbf{e}_i^j}^D||^2 \,,
\label{eq:motionObj2}
\end{align}
where $k$ is the number of end-effectors, $\mathbf{x}_i^D$ and $\mathbf{e}_i^D$ represent desired robot COM, and end-effector positions respectively. Apart from these, motion smoothness is often desired by the users, which is encoded by the following objective:
\begin{align}
E_{\text{Smooth}} = \frac{1}{2}\sum_{i=2}^{T-1}|| \mathbf{q}_{i-1} - 2 \mathbf{q}_i + \mathbf{q}_{i+1} ||^2 \,.
\label{eq:motionObj3}
\end{align}

\subsubsection{Constraints}
We next define various constraints to ensure that the generated motion is stable. 

\textbf{Kinematic constraints:} The first set of constraints ask the position of COM, and end-effectors to match with the pose of the robot. For every time step $i$, and end-effector $j$:
\begin{align}
\varphi_{COM}(\mathbf{q}_i) - \mathbf{x}_i &= 0\,, \nonumber \\
\varphi_{EE}(\mathbf{q}_i)^j - \mathbf{e}_i^j &= 0\,, \quad \forall j, 
\label{eq:constrain1_2}
\end{align}
where $\varphi_{COM}$ and $\varphi_{EE}$ are forward kinematics functions outputting the position of COM and end-effectors respectively.

We also have a set of constraints that relate the net force and torque to the acceleration and angular acceleration of the robot's COM:
\begin{align}
\sum_{j=1}^k c_i^j\mathbf{f}_i^j &= M\ddot{\mathbf{x}}_i\,, \nonumber \\
\sum_{j=1}^k c_i^j(\mathbf{e}_i^j - \mathbf{x}_i^j)\times\mathbf{f}_i^j &= \mathbf{I}\ddot{\mathbf{o}}_i\,, 
\label{eq:constrain3_4}
\end{align}
where $M$ is the total mass of the robot, and $\mathbf{I}$ is the moment of inertia tensor. The acceleration $\ddot{\mathbf{x}}_i$ can be evaluated using finite differences: $\ddot{\mathbf{x}}_i = (\mathbf{x}_{i-1} - 2\mathbf{x}_i + \mathbf{x}_{i+1})/h^2$, where $h$ is the time step. Similarly, the angular acceleration $\ddot{\mathbf{o}}_i$ can be expressed as $ \ddot{\mathbf{o}}_i = (\mathbf{o}_{i-1} - 2\mathbf{o}_i + \mathbf{o}_{i+1})/h^2$. We note that the orientation of the root $\mathbf{o}_i$ is part of the pose $\mathbf{q_i}$, and it uses axis-angle representation.

\textbf{Friction constraints:} To avoid foot-slipping, we also have the following constraints for each end-effector $j$:
\begin{align}
c_i^j(\mathbf{e}_{i-1}^j - \mathbf{e}_{i}^j) = 0, \, c_i^j(\mathbf{e}_{i}^j - \mathbf{e}_{i+1}^j) = 0\,,
\label{eq:constrain5}
\end{align}
for all $2\leq i \leq T-1$, which implies that the end-effectors are only allowed to move when they are not in contact with the ground. Further, to account for different ground surfaces, we enforce the friction cone constraints:
\begin{align}
{f_i^j}_\parallel \leq \mu {f_i^j}_\bot \,,
\label{eq:constrain7}
\end{align}
where ${f_i^j}_\parallel$ and ${f_i^j}_\bot$ denote the tangential and normal component of $\mathbf{f}_i^j$ respectively, and $\mu$ is the coefficient of friction of the ground surface.

\textbf{Limb collisions:} For physical feasibility, we propose a collision constraint that ensures a safe minimum distance between the limb segments of robot over the entire duration of the motion. 
\begin{align}
d(\mathbf{V}_i^{k_1}, \mathbf{V}_i^{k_2}) \leq \delta\,,
\label{eq:constrain8}
\end{align}
where $\mathbf{V}_i^{k}$ represents a 3D segment representing the position and orientation of $k^{th}$ limb, $d(\cdot)$ computes the distance between $k_1$ and $k_2$ limbs, and $\delta$ is the threshold distance beyond which collisions may happen.

\textbf{Motion periodicity:} If the users prefer a periodic motion, we can add an additional constraint that relates the start pose $\mathbf{q}_1$ and the end pose $\mathbf{q}_T$ of the robot:
\begin{align}
\mathbf{J}(\mathbf{q}_T) - \mathbf{J}(\mathbf{q}_1) = 0\,,
\label{eq:constrain6}
\end{align}
where $\mathbf{J}(\mathbf{q_i})$ extract the orientation of the root and joint parameters from pose $\mathbf{q_i}$.

Finally, $F(\mathbf{s},\mathbf{m})$ is computed as the sum of the objectives in eq.~\ref{eq:motionObj1},~\ref{eq:motionObj2},~\ref{eq:motionObj3}. Quadratic penalty costs for violation of constraints in eq.~\ref{eq:constrain1_2},~\ref{eq:constrain3_4},~\ref{eq:constrain5}, and~\ref{eq:constrain6} are also added to $F$. Note that, while these objective terms are defined using the motion parameters $\mathbf{m}$, $F$ is indirectly affected by the structural parameters $\mathbf{s}$ as discussed in Sec.~\ref{sec:manifold}.


\section{Results}
\label{sec:results}

\begin{figure}[htbp]
	\centering
	\includegraphics[width = \columnwidth]{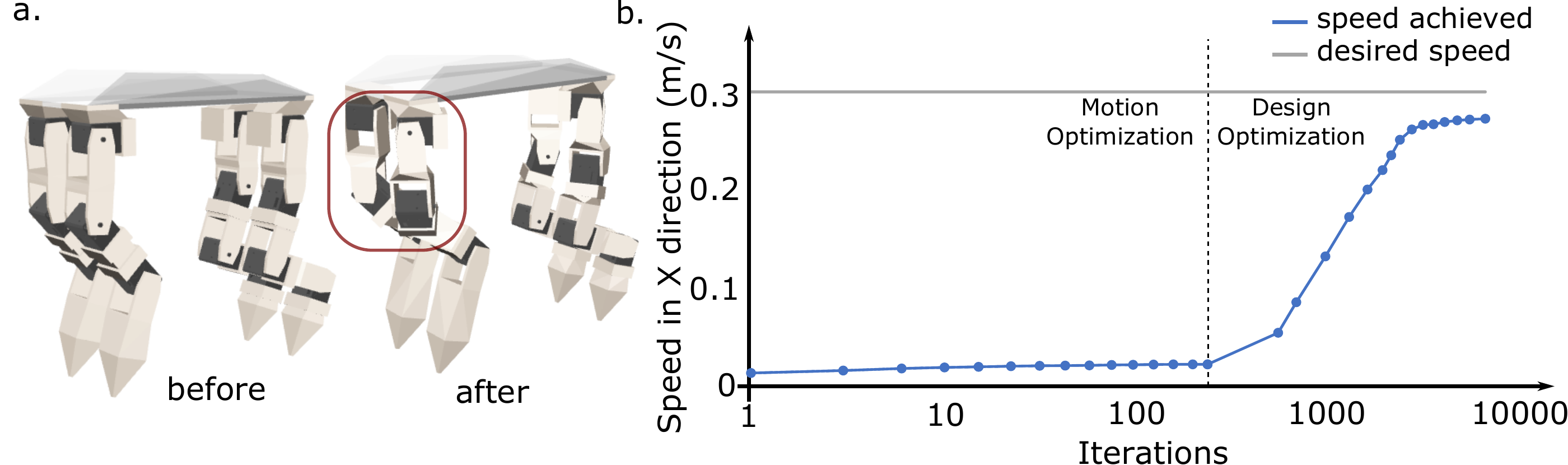}
	\caption{(a)~Initial and optimized designs of a `puppy' robot are shown. The initial design can only walk forward (in Z direction) owing to the deficient placement of actuators at its joints. Our design optimization changes the actuator orientations to enable the robot to walk sideways (in X direction).~(b)~Optimizing motion parameters is not enough for overcoming such design deficiencies as seen in the initial iterations of the plot. On the other hand, optimization of the structure parameters modifies the design to walk sideways with desired speed.}
	\label{result1}
\end{figure}

We explore three simulated examples to study the utility and effectiveness of our approach. Although, we only show our current findings in simulation, we have confirmed that the simulation environment matches physical results, in our past work~\cite{desai2017computational}. The first example is aimed to show how our design optimization can change an initial robot design that is inadequate to perform a user-specified task. The second example shows how our system ensures feasibility while maintaining task performance. Finally, we show how our system can be used to optimize designs for multiple tasks. We also demonstrate how different tasks may require varied design changes, and how our optimization can automatically generate these  design changes in matters of minutes, thereby illustrating its computational efficiency. Note that all the initial designs for these examples were manually designed with our interface. Our video\footnote{Video is available at: \url{https://tinyurl.com/RoboCodesign}} demonstrates this process for the robot in fig.~\ref{result1}(a). 


When novices design robots, it can be hard for them to decide where the actuators should be located and how they should be oriented for achieving a specific behavior. Figure~\ref{result1}(a) shows one such example of a `puppy' robot with three motors per leg. Even with enough number of actuators, the robot can only walk in one direction (forward) owing to its actuator placements. In particular, all actuators rotate about the same axis thereby reducing the effective degrees of freedom (DOF) of each limb. By parameterizing the actuator orientations $\mathbf{a_i}$ in eq.~\ref{eq:robot}, we enable our design optimization to change them for equipping the robot to walk in any specific direction. Without the optimization of such structural parameters, it may be impossible to achieve such tasks. For instance, the optimization of only motion parameters for the initial `puppy' design fails to produce the desired behavior of walking sideways at a specific speed (see fig.~\ref{result1}(b)).

\begin{figure}[htbp]
	\centering
	\includegraphics[width = \columnwidth]{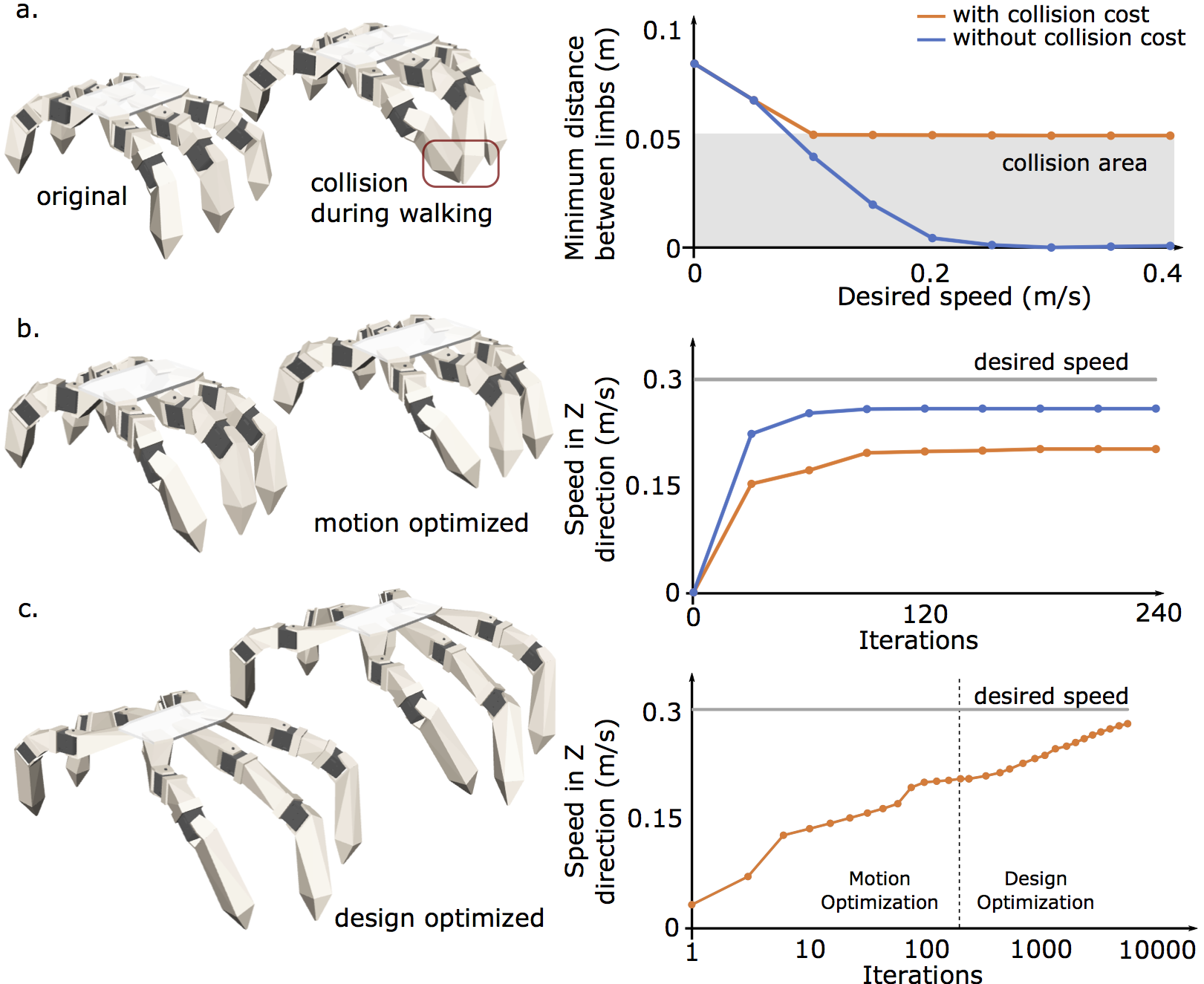}
	\caption{(a)~A hexapod robot's limbs start colliding at higher speed. Accounting for collisions during motion optimization prevents this, but also restricts the robot from walking faster as seen in the plots in~(a), and~(b).~(c)~Design optimization is able to deal with this trade-off by increasing the spacing between limbs, and their lengths. This enables the robot to walk faster without any collisions.}
	\label{result2}
\end{figure}

Even when a design can theoretically achieve the desired behavior, it may be rendered infeasible due to real world constraints such as collisions. Fig.~\ref{result2}(a) shows a robot that can walk in the user-specified direction at desired speeds. However, when walking speeds increase above 0.1 $ms^{-1}$, the robot's limbs start colliding. It is hard to anticipate such issues apriori. Along with helping the user to test such scenarios in simulation, our system can also automatically prevent them by using feasibility constraints during motion optimization as discussed in Sec.~\ref{sec:mopt}. However, these constraints prevent the required range of limb motions needed for fast walking, limiting the ability of the robot to perform the intended task (see fig.~\ref{result2}(b)). On the other hand, design optimization is able to change the design to achieve both these contradicting tasks successfully as seen in fig.~\ref{result2}(c).

\begin{figure}[htbp]
	\centering
	\includegraphics[width = \columnwidth]{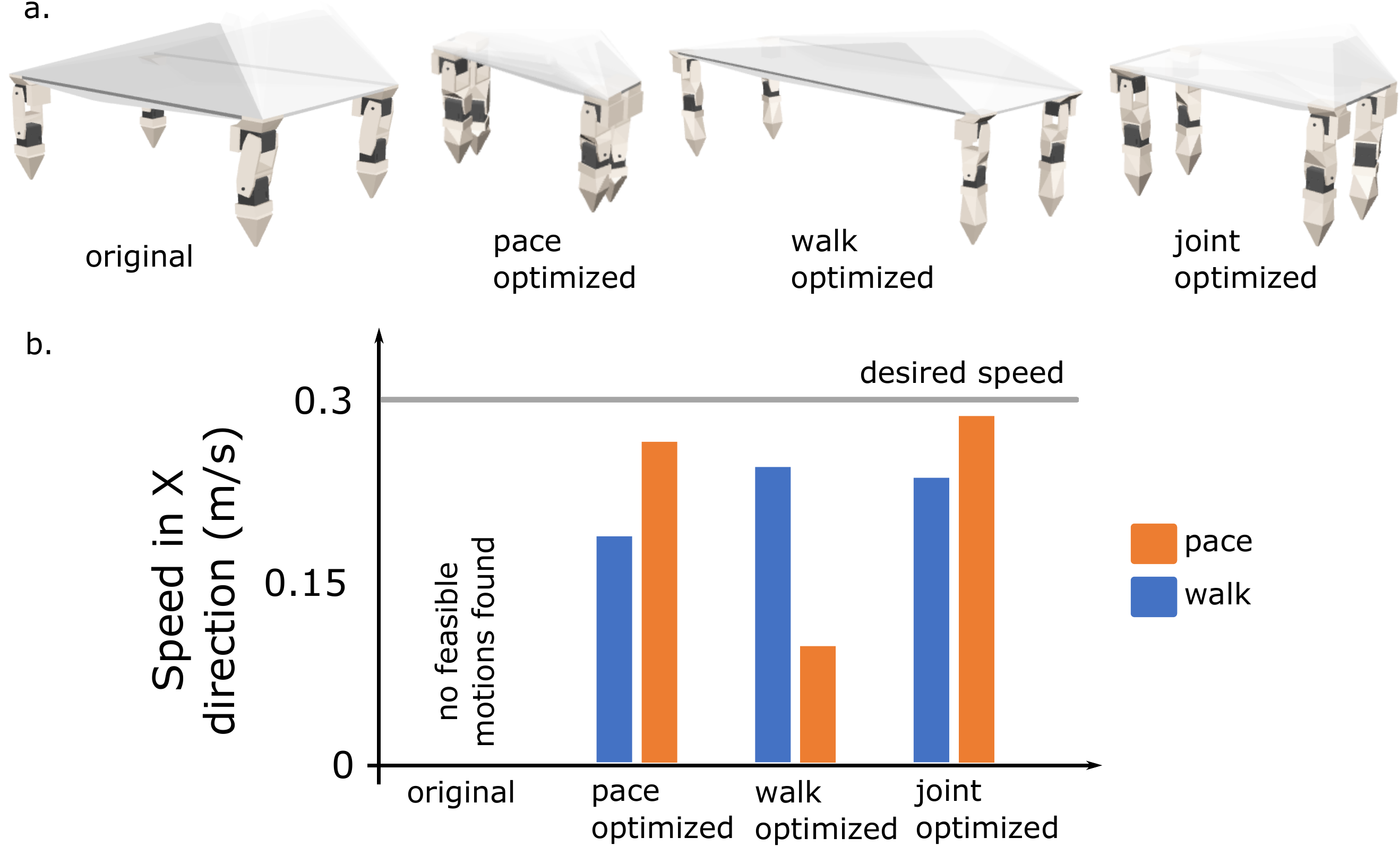}
	\caption{(a)~An initial quadruped robot design that can only walk forward, is optimized to pace and walk sideways respectively. A design that was jointly optimized for both these behaviors simultaneously is also shown. Note that the body aesthetics was manually added to each design ~(b)~Unlike designs generated by optimization that considered the tasks of walking and pacing separately, jointly optimized design achieves a reasonable trade-off between the performance of both tasks.}
	\label{result3}
\end{figure}

Finally, designing robots for multiple tasks is also highly challenging, especially if the tasks ask for opposing design characteristics. Consider the task of walking and pacing for a quadruped robot shown in fig.~\ref{result3}(a). The original design can only walk forward owing to its actuator placements (similar to the `puppy' robot in fig.~\ref{result1}(a)). Its wider body and shorter limbs prevent it from pacing in stable manner. The tasks are further challenging because of limited number of actuators. While three actuators may enable sufficient DOF for 3D movements without careful placement of individual actuators, same may not be true for designs with lower number of actuators. Individually optimizing the design for pacing and walking may not be sufficient for enabling the robot to perform both tasks. Pace based design optimization generates a slim bodied robot, while walk based design optimization produces a wider body size to increase stability during fast walking (see fig.~\ref{result3}(a)). Such a wider body in turn, negatively affects the pacing behavior (fig.~\ref{result3}(b)). To achieve reasonable performance for both these tasks, a trade-off is thus required. Our system allows users to jointly optimize their designs for multiple tasks, to handle such scenarios. The individual requirements for each task $F_i(\mathbf{s},\mathbf{m_i})$ can be combined in weighted manner into $F_{joint}(\mathbf{s},\mathbf{m)} = \sum w_i F_i(\mathbf{s},\mathbf{m_i})$. Weights $w_i$ representing the importance of each task can be set by the users. Such joint optimization of walking and pacing (with $w_1 = w_2 = 0.5$) for quadruped in fig.~\ref{result3}(a) succeeds in achieving the necessary trade-off as illustrated in the resultant medium bodied optimized design, and the corresponding task performance.

\setlength{\tabcolsep}{0.95mm}
 \begin{table}[hbtp!]{
		\begin{center}
			\begin{tabularx}{\columnwidth}{c|c c c c }
				Robot & Number of & Motion Opt.&  Design Opt.& Time \\
				& Parameters & Iterations & Iterations & (s)\\
				\cline{1-5}
				Puppy (Fig.~\ref{result1})&614&6207&32&107.66\\
				Hexapod (Fig.~\ref{result2})&1044&5013&53&97.47\\
				Quadruped (Fig.~\ref{result3})&1050&14873&100&124.47\\
				\cline{1-5}
			\end{tabularx}
		\end{center}
		\caption{Design optimization statistics for example robots}
		\label{tab:results}
	}
\end{table}

Table~\ref{tab:results} shows the design times for optimizing the designs of robots in fig.~\ref{result1},~\ref{result2},~\ref{result3}. For quadruped in fig.~\ref{result3} these statistics are reported for the joint optimization scenario. Note that, even when its number of optimization parameters are roughly similar to that of the hexapod, there is a significant difference in the number of optimization iterations, and the time required. This is because of the contradicting requirements that the two tasks demand, making the problem more challenging. Also note that for each iteration of design optimization, multiple iterations of motion optimization are executed (see Algorithm~\ref{algo}). However, as illustrated in the statistics, the large number of these iterations needed to update the designs are executed in matter of minutes. Such computational efficiency is at the core of interactivity in our system. Apart from an efficient implementation in $\CC$, a scalable approach using the Adjoint method allows us to achieve the same. While it is hard to make a direct comparison owing to differences in parameterization and implementation, our design times are significantly faster than the current state-of-the-art that co-optimizes form and function for many such similar robots~\cite{spielberg2017functional}. For instance, their system took on an average 685 seconds to optimize a biped robot design for the task of walking on flat ground (750 optimization parameters)~\cite{spielberg2017functional}.

\section{Discussion and Future Works}
\label{sec:discussion}
We introduced an interactive robot design and optimization system that allows casual users to create customized robotic creatures for specific tasks. Apart from generating feasible behaviors, our system improves the performance of robot through an automatic design optimization process. This is particularly important for users who do not have a strong background in the relevant areas. Assisted by our system, they can eliminate the deficiencies of their design for the task at hand. Further, modern robots are increasingly demanded to have a variety of functionalities. We therefore also introduced the feature of joint optimization of the robot designs for multiple behaviors.

By introducing an easy-to-use design system with high user interactivity, we made a step forward in increasing the accessibility of robotics. However, some limitations should be noticed. For this work, we focused our efforts on periodic locomotion-based tasks. In the future, we plan to extend our design optimization technique for a broader class of motions and behaviors, including climbing, carrying weights or avoiding obstacles. Moreover, our system does not perceive the mass or the number of motors as design parameters. However, our approach is generic, in that the adjoint method can be applied to any trajectory optimization scheme that provides analytical gradients and Hessians. Therefore, although our motion optimization currently does not account for the mass of the limbs, we believe that this limitation can be eliminated in the future. Finally, the optimization processes are relatively invisible to the user in our system, which provides little room of manipulation for experts and specialists who may desire finer control over the design process. We therefore believe that it is important
to find the right balance between automation and user control during design. Towards better understanding this balance for novice, intermediate, and experienced robot designers, we plan to perform an
extensive user study as well in the future.




\bibliographystyle{IEEEtran}
\bibliography{library}

\end{document}